\documentclass{article}

\PassOptionsToPackage{numbers, compress}{natbib}
 \usepackage[preprint]{neurips_2026}

\usepackage[utf8]{inputenc} 
\usepackage[T1]{fontenc}    
\usepackage{hyperref}       
\usepackage{url}            
\usepackage{booktabs}       
\usepackage{amsfonts}       
\usepackage{nicefrac}       
\usepackage{microtype}      
\usepackage{xcolor}         

\usepackage{booktabs} 
\usepackage{pifont}   
\usepackage{amsmath}
\usepackage{graphicx}
\usepackage{subcaption}  
\usepackage{wrapfig}
\usepackage{multirow}
\usepackage{tcolorbox}
\usepackage[table]{xcolor}

\newcommand{\cmark}{\textcolor{green}{\ding{51}}}   
\newcommand{\xmark}{\textcolor{red}{\ding{55}}}     

\title{Omni-DuplexEval: Evaluating Real-time Duplex Omni-modal Interaction}

\author{%
\begin{tabular}{c}
Chaoqun He$^{1}$\thanks{Equal contribution. Emails: Chaoqun He (hecq25@mails.tsinghua.edu.cn)} \quad
Mingyang Xiang$^{2}$\footnotemark[1] \quad
Yingjing Xu$^{3}$ \quad
Bokai Xu$^{3}$ \quad
Junbo Cui$^{3}$ \\
Jie Zhou$^{3}$ \quad
Yuan Yao$^{1}$\thanks{Corresponding authors.} \quad
Lijie Wen$^{1}$\footnotemark[2] \\
{\normalfont
$^{1}$Tsinghua University \quad
$^{2}$Shanghai Qi Zhi Institute \quad
$^{3}$ModelBest Inc.} \\
\\
{\normalfont \textbf{Project:} \url{https://github.com/OpenBMB/Omni-DuplexEval}}
\end{tabular}
}

\newcommand{\name}[0]{Omni-DuplexEval}

\begin{document}

\maketitle

\begin{abstract}
Real-time duplex interaction is essential for multimodal AI systems operating in real-world scenarios, where models must continuously process streaming inputs and respond at appropriate moments. However, most existing multimodal large language models (MLLMs) are evaluated in offline settings, where the entire video input is processed before any response is generated. While recent work has started to explore real-time duplex MLLMs, there is still no comprehensive benchmark or automatic evaluation method for this setting. To address this gap, we propose \textbf{\name{}}, \textbf{a benchmark for systematically evaluating real-time duplex interaction}. The benchmark consists of two complementary scenarios:
(1) \textbf{Real-Time Description}, which evaluates the ability to generate continuous, time-aligned responses that track evolving multimodal inputs, and
(2) \textbf{Proactive Reminder}, which evaluates the ability to identify salient events and respond at appropriate moments. 
\name{} contains 660 videos with fine-grained, human-annotated labels and precise temporal metadata, spanning 9 tasks grounded in real-world scenarios, where all questions are formulated as open-ended queries. We further introduce an automatic evaluation framework based on LLM-as-a-Judge, which enables systematic assessment by jointly evaluating response–content alignment and response timing through timestamp-aware and sequential reasoning, achieving strong alignment with human judgments.
Experiments on state-of-the-art duplex MLLMs reveal substantial limitations. 
The best-performing model achieves only 39.6\% overall, while scoring only 20.0\% on Proactive Reminder.
Our analysis identifies two key challenges: \textbf{models struggle to balance timely responses with coherent, holistic content generation, and they often fail to determine both when to respond and what to produce.} 
We hope our work facilitates further progress in MLLMs, particularly in real-time duplex interaction.
\end{abstract}

\section{Introduction}
Multimodal Large Language Models (MLLMs) have achieved strong performance on video understanding task, with recent systems such as GPT-4o~\citep{hurst2024gpt} and Gemini-Pro~\citep{gemini31pro_modelcard} demonstrating impressive capabilities. 
However, most of existing models are designed for static images or offline video processing and must observe the entire video before producing a response. This setting is commonly used in current benchmarks, such as Video-MME~\citep{fu2025video}, LVBench~\citep{wang2025lvbench}.
This offline setting differs fundamentally from real-world interaction, where perception and response are tightly coupled: humans observe, listen, and respond simultaneously~\citep{lin2025full}, enabling continuous and real-time interaction without waiting for complete information.
We refer to this capability as \textbf{real-time duplex interaction}, where models process continuously evolving inputs and produce responses at appropriate moments.

Recent advances have begun to explore streaming MLLMs that can process inputs and generate outputs incrementally. Systems such as LiveCC~\citep{chen2025livecc} demonstrate the ability to produce real-time video commentary, while MiniCPM-o 4.5~\citep{yao2024minicpm} supports full-duplex multimodal live streaming. These systems exhibit early forms of real-time duplex behavior.

\begin{figure}[htbp]       
    \centering            
    \includegraphics[width=\textwidth]{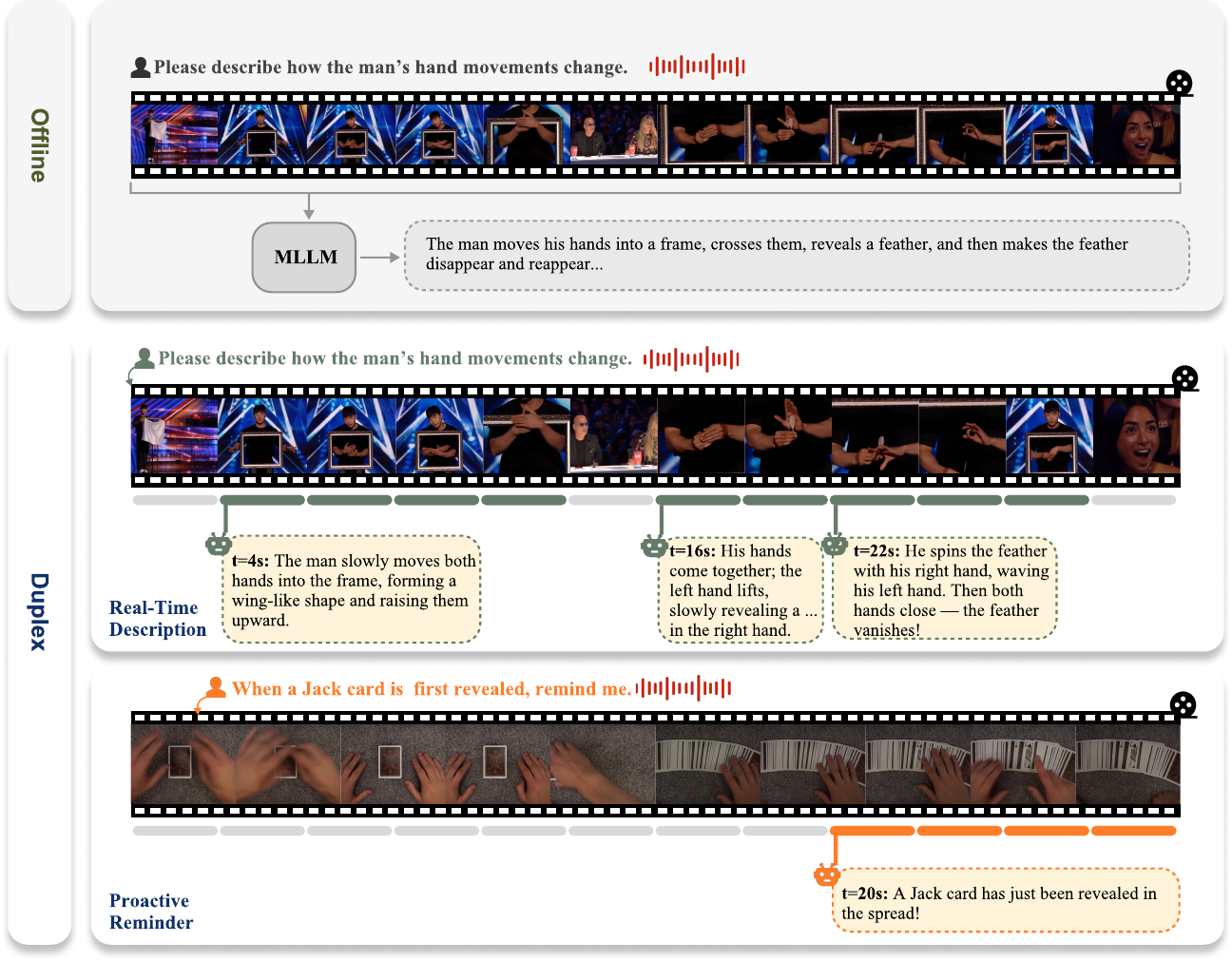}
    \caption{Comparison between \name{} and offline evaluation paradigms. Offline settings require models to process the entire video before producing a response. In contrast, \name{} introduces two scenarios to evaluate real-time duplex capabilities, including continuous response generation over evolving video content and the ability to determine when to respond and what to say.
}   
    \label{fig:main_fig}   
\end{figure}

However, current benchmarks for video understanding do not fully capture these capabilities. For example, StreamingBench~\citep{lin2026streamingbench} and OVOBench~\citep{li2025ovobench} primarily rely on multiple-choice formats and focus on final response quality, without capturing temporal alignment or continuous adaptation. OmniMMI~\citep{wang2025omnimmi} provides open-ended responses, but its answers are relatively simple and sparse, making it difficult to assess response quality in realistic settings. ProactiveVideoQA~\citep{wang2025proactive} and PhoStream~\citep{lu2026phostream} focus on proactivate detection and interaction, but lack fine-grained evaluation of temporal dynamics and response behavior over time. As a result, current benchmarks do not adequately evaluate real-time duplex capabilities.

To address this gap, we introduce \textbf{\name}, a benchmark designed to evaluate real-time duplex capabilities, where models are expected to process evolving video inputs and produce responses at appropriate moments.
The benchmark is organized into two complementary scenarios as shown in Figure~\ref{fig:main_fig}.
Real-Time Description evaluates the ability to process evolving video inputs and generate responses continuously while adapting to changes in the video.
Proactive Reminder evaluates the ability to detect relevant events and determine when to respond, producing appropriate outputs in response to user instructions grounded in the video. The benchmark includes 660 samples, each paired with an open-ended question and detailed human annotations. It covers 9 tasks designed to reflect real-world scenarios, spanning diverse domains such as entertainment, lifestyle, and education.

Furthermore, existing evaluation approaches are not well suited for assessing real-time duplex capabilities. To address this, we propose an automatic evaluation framework based on LLM-as-a-Judge. The framework jointly evaluates semantic correctness and response timing, enabling flexible assessment of both what to say and when to say it. This provides a practical way to measure real-time duplex behavior beyond traditional final-answer-based evaluation.

We conduct extensive experiments on recent duplex omni-modal models. Results expose two fundamental gaps. In Real-time Description, models exhibit a completeness-timeliness trade-off, remaining silent for approximately 50-60\% of the video duration and failing to provide continuous description. 
In Proactive Reminder, models struggle not with what to say but with when to say it. In most cases, models fail to produce responses at the appropriate time, often remaining silent. As a result, performance is consistently low, with the best model achieving only 20.0\%.
These findings suggest that current models remain far from supporting real-world interactive assistants.
We hope that \textbf{\name{}} will facilitate future research on real-time duplex omni-modal interaction.


\begin{table}[htbp]
\caption{Comparison of \textbf{\name} with other representative video and audio-visual benchmarks. \textbf{V} = Visual, \textbf{A} = Audio, \textbf{Sub} = Subtitles, \textbf{I} = Image. \textbf{Open-Ended} denotes whether the benchmark evaluates free-form textual responses rather than multiple-choice questions. \textbf{Streaming} indicates the ability to handle sequential video inputs. \textbf{Proactive} evaluates whether the system can autonomously determine response timing without user queries. \textbf{Temporal Alignment} assesses the physical synchronization between streaming inputs and generated texts.}
\label{tab:benchmark_comparison}
\centering

\resizebox{\textwidth}{!}{
\begin{tabular}{l c c c c c c}
\toprule
\textbf{Benchmark} & \textbf{Modality} & \textbf{\#Videos} & \textbf{Open-Ended} & \textbf{Streaming} & \textbf{Proactive} & \textbf{Temporal Alignment} \\ 
\midrule
\multicolumn{7}{l}{\textit{Offline Benchmarks}} \\
MVBench~\citep{li2024mvbench} & V & 3,641 & \xmark & \xmark & \xmark & \xmark \\
Video-MME~\citep{fu2025video} & V, Sub & 900 & \xmark & \xmark  & \xmark & \xmark \\
MLVU~\citep{zhou2024mlvu} & V & 1,730 & \cmark & \xmark & \xmark & \xmark \\
LongVideoBench~\citep{wu2024longvideo} & V & 3,763 & \xmark & \xmark & \xmark & \xmark \\
OmniBench~\citep{li2026omnibench} & A, I & - & \xmark & \xmark & \xmark & \xmark \\
WorldSense~\citep{hong2025worldsense} & V, A & 1,662 & \xmark & \xmark & \xmark & \xmark \\
\midrule
\multicolumn{7}{l}{\textit{Online Benchmarks}} \\
StreamingBench~\citep{lin2026streamingbench} & V, A & 900 & \xmark & \cmark & \cmark & \xmark \\
OVOBench~\citep{li2025ovobench} & V & 644 & \cmark & \cmark & \cmark & \xmark \\
OmniMMI~\citep{wang2025omnimmi} & V, A & 1,121 & \cmark & \cmark & \cmark & \xmark \\
ProactiveVideoQA~\citep{wang2025proactive} & V, A & 1,377 & \cmark & \xmark & \cmark & \xmark \\
PhoStream~\citep{lu2026phostream} & V, A & 578 & \cmark & \cmark & \xmark & \xmark \\
RIVER~\citep{shi2026river} & V & 1,067 & \cmark & \cmark & \cmark & \xmark \\
\midrule
\textbf{\name} & V, A & 660 & \textbf{\cmark} & \textbf{\cmark} & \textbf{\cmark} & \textbf{\cmark} \\
\bottomrule
\end{tabular}
}
\end{table}

\section{Related Works}

\subsection{Video MLLM}

Multimodal Large Language Models (MLLMs) have evolved from early video understanding systems that rely on auxiliary signals to unified architectures integrating visual, audio, and textual information~\citep{li2023videochat, su2023pandagpt, chen2023vast}. Recent "omni-modal" models aim to uniformly process multiple modalities within a single architecture~\citep{wu2023next, han2024onellm, fu2024vita}. Efficient MLLM designs have also emerged, achieving strong performance with fewer parameters through adaptive visual encoding~\citep{yao2024minicpm, hong2024cogvlm2}.

Despite these advances, most existing MLLMs operate under an offline paradigm. To address this, recent streaming models process inputs incrementally and support streaming generation, moving toward full-duplex multimodal interaction~\citep{chen2024videollm, zhang2025flash, xu2025streamingvlm, chen2025livecc, wang2026streambridge, sun2025video}. Recent advances have also introduced scene-aware optimization for efficient long-context reasoning in streaming QA, as well as unified evaluation protocols that characterize trade-offs between efficiency, storage, and accuracy under realistic constraints~\citep{lu2026vista, tang2026streamingeval}.

\subsection{Evaluation Benchmarks}

Traditional offline video understanding benchmarks have evolved from short-video perception to complex reasoning and long-form comprehension, covering multi-task evaluation and long video understanding~\citep{li2024mvbench, fu2025video, zhou2024mlvu, wu2024longvideo, li2026omnibench, hong2025worldsense}. Specialized benchmarks have also been developed for ego-centric and activity understanding~\citep{mangalam2023egoschema, patraucean2023perception, yu2019activitynet}. A comprehensive survey systematically analyzes the landscape of VideoLLM benchmarks and evaluation methodologies~\citep{kumar2025videollm}.

Recent benchmarks have begun exploring streaming and real-time evaluation. Early efforts introduce streaming settings but largely rely on multiple-choice formats and focus on final response quality~\citep{lin2026streamingbench, li2025ovobench, xun2026rtv, chen2025livecc}. Subsequent work moves toward interactive and proactive evaluation, incorporating event-driven tasks and proactive reasoning into streaming video understanding~\citep{wang2025omnimmi, lu2026phostream, shi2026river, wang2025proactive}. More recent benchmarks propose continuous evaluation metrics and standardized protocols for assessing proactiveness and temporal consistency~\citep{chatterjee2026don, vasu2026vsas, tao2026lvomnibench}.

Beyond streaming settings, new benchmarks have been established for omni-modal understanding, evaluating multimodal reasoning on large-scale real-world videos with questions requiring tight coupling of visual and audio signals~\citep{goel2026mmou, xie2026maverix}. For hallucination evaluation, recent work systematically defines multiple types of video QA hallucinations and constructs multi-round open-ended benchmarks~\citep{wildvideo2025}. For full-duplex spoken interaction, benchmarks have been proposed to evaluate turn-taking capabilities and handle real-time interruptions and overlapping speech~\citep{lin2025full, wang2026full}.

Despite these advances, existing benchmarks do not comprehensively evaluate real-time duplex interaction—the ability to generate continuous responses while maintaining temporal alignment with evolving video streams. They largely focus on discrete question-answering rather than continuous streaming generation, and treat response timing separately from content correctness. Our \name{} addresses these limitations through unified evaluation of what to say and when to say it.
Table~\ref{tab:benchmark_comparison} presents a comparison between our benchmark and other representative benchmarks.

\section{\name}
\label{sec:datasets}

\subsection{Taxonomy}

Real-time duplex capability requires models to process continuously evolving inputs and produce responses at appropriate moments.
Based on this, \name{} is organized into two representative scenarios.
Real-Time Description evaluates the ability to generate responses that follow evolving video content in real time. Proactive Reminder evaluates the ability to identify relevant events and determine when to respond.
We describe these two scenarios in detail below.


\subsubsection{Real-Time Description}
Real-Time Description evaluates the ability to generate responses that follow evolving video content in real time. At the beginning of each sample, the model receives a user instruction that specifies a particular subject or aspect of interest, and produces continuous, time-aligned responses as the video unfolds.
The responses should remain grounded in the instruction while reflecting changes in the current temporal window, requiring the model to track dynamic visual and auditory information and update its outputs accordingly.

\begin{figure}[htbp]       
    \centering            
    \includegraphics[width=1\textwidth]{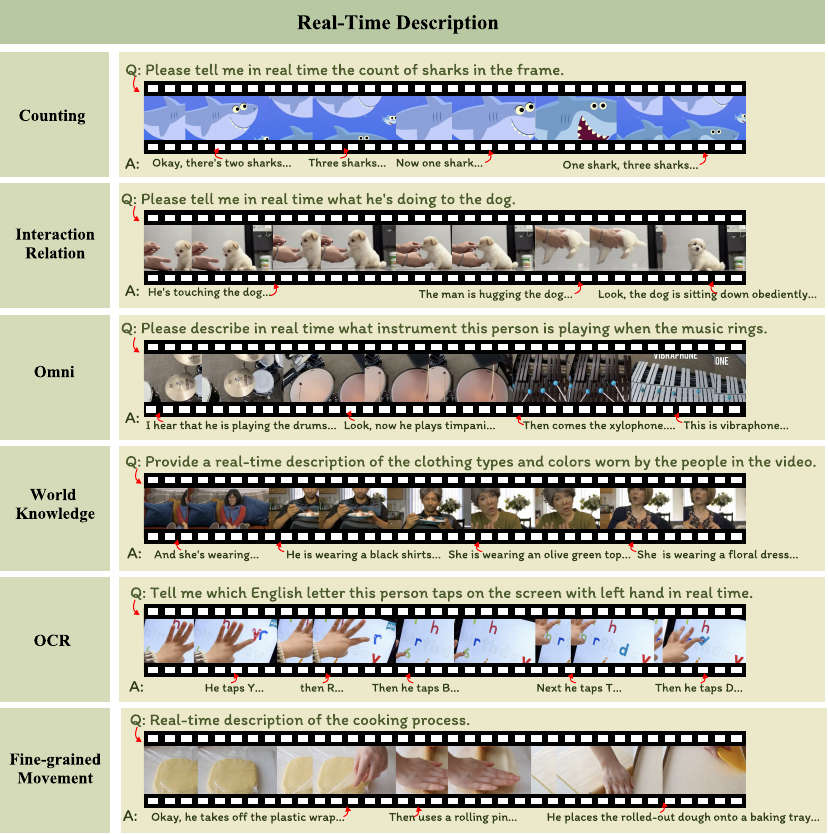}
    \caption{Example of each task in Real-Time Description.}   
    \label{fig:realtime_description}   
\end{figure}

To evaluate this capability, we define six sub-tasks within the Real-Time Description as shown in Figure~\ref{fig:realtime_description}. 
(1) \textbf{Counting} (CT) assesses the model’s capacity for incremental tallying and temporal consistency as it tracks the entry, exit, or occlusion of objects (e.g., fluctuating pedestrian counts) in a fluid scene. 
(2) \textbf{Interaction Relation} (IR) examines the model’s understanding of the social or physical connections between multiple entities. It requires describing how people or objects interact as those relationships unfold dynamically. 
(3) \textbf{Omni}, as the most comprehensive task, Omni requires the model to synthesize both visual and auditory streams simultaneously. 
(4) \textbf{World Knowledge} (WK) evaluates the model's ability to identify specific attributes and categories—such as animal species, clothing materials, or commercial brands. 
(5) \textbf{OCR} focuses on dynamic text perception, this task requires the model to recognize and read out characters that evolve over time, such as scrolling subtitles or changing floor numbers in an elevator, demanding precise synchronization between visual transitions and textual output. 
(6) \textbf{Fine-grained Movement} (FM) focuses on capturing high-fidelity trajectories of complex movements, translating granular biological or mechanical actions (e.g., intricate hand gestures) into precise descriptors via short-term temporal dependencies. 

\begin{figure}[htbp]       
    \centering            
    \includegraphics[width=1\textwidth]{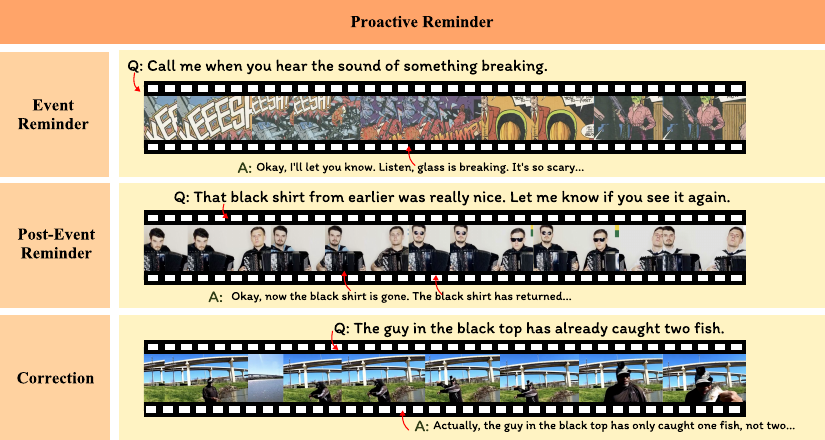}
    \caption{Example of each task in Proactive Reminder.}   
    \label{fig:proactive_reminder}   
\end{figure}

\subsubsection{Proactive Reminder}

Proactive Reminder evaluates the ability to identify relevant events and determine when to respond based on streaming video inputs. The model receives a user instruction that specifies a clear and well-defined event, and must monitor the incoming omni-modal stream to produce a response when the event occurs. This requires the model to retain the instruction, track visual and auditory information over time, and decide both when to respond and what to say. In some cases, the instruction may appear at arbitrary points along the video timeline, requiring the model to relate it to past observations.

We further divide this scenario into three sub-tasks as shown in Figure~\ref{fig:proactive_reminder}:
(1) \textbf{Event Reminder} (ER). The instruction describes a future event. The model monitors the video stream and produces a response when the event occurs.
(2) \textbf{Post-Event Reminder} (PER). The instruction refers to a past event. The model determines whether the event occurs again and produces a response accordingly.
(3) \textbf{Correction} (CR). The instruction contains an incorrect description of the video. The model is expected to revise the description based on the observed content.

Together, these two scenarios capture both continuous and event-driven response patterns in real-time settings, providing complementary evaluation of real-time duplex interaction capabilities.
They also place strong demands on omni-modal perception and reasoning, requiring models to effectively integrate visual and auditory signals and perform real-time analysis.
\subsection{Benchmark Construction}

After defining the task taxonomy, we construct the dataset to reflect general real-time duplex interaction scenarios. Videos are collected from diverse online sources and filtered to ensure quality and diversity. We retain videos with clear temporal dynamics and omni-modal signals (e.g., visual and auditory changes), while removing static or low-information content. This design ensures that the dataset emphasizes time-evolving interactions rather than static scene understanding.

To support reliable evaluation, we carefully design question–answer pairs for each scenario. For Real-Time Description, we identify a subject with continuous temporal variation in each video and construct questions that require describing its evolving state, rather than providing generic summaries. This encourages models to focus on specific entities and track their changes over time, aligning with real-world interaction patterns. Annotators generate responses by continuously observing the video and describing these changes in real time. Each sample is annotated by two independent annotators, with a third annotator resolving disagreements to ensure annotation consistency. For Proactive Reminder, questions are introduced at arbitrary points along the video timeline to simulate real-time user interaction. Each question specifies a clear and unambiguous event, and ground-truth annotations are aligned with the corresponding event timestamps. 
In the Proactive Reminder scenario, some samples contain multiple occurrences of the target event, requiring models to handle repeated event detection and response.

Finally, all samples undergo strict quality control, including cross-annotation consistency checks and validation of temporal annotations, ensuring the reliability of the dataset.

\begin{figure}[t]
    \centering

    \begin{subfigure}[b]{0.34\linewidth}
        \centering
        \includegraphics[width=\linewidth]{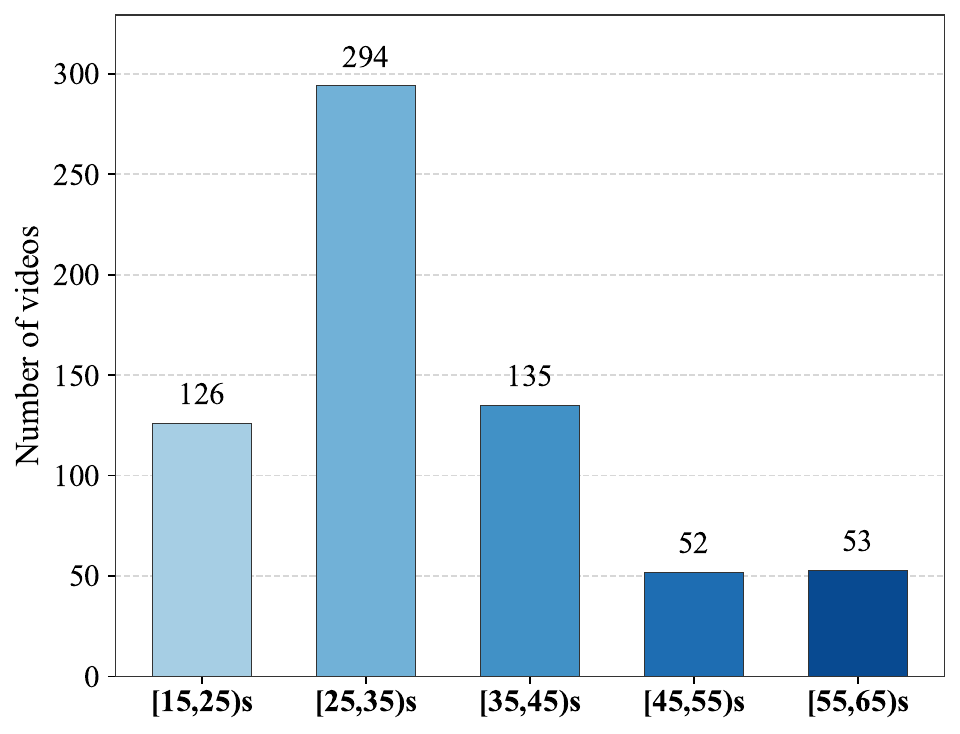}
    \end{subfigure}
    \hfill
    \begin{subfigure}[b]{0.28\linewidth}
        \centering
        \includegraphics[width=\linewidth]{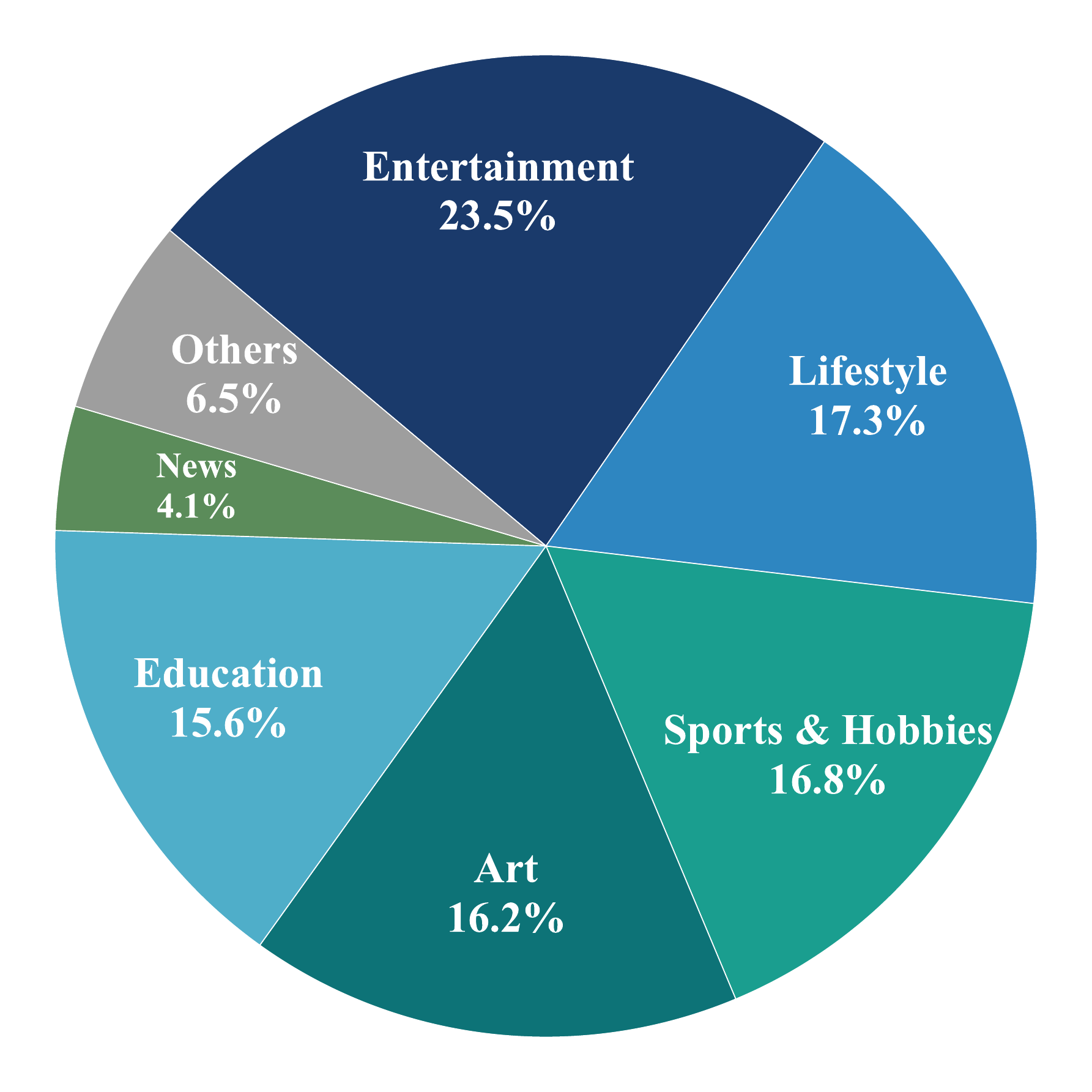}
    \end{subfigure}
    \hfill
    \begin{subfigure}[b]{0.35\linewidth}
        \centering
        \includegraphics[width=\linewidth]{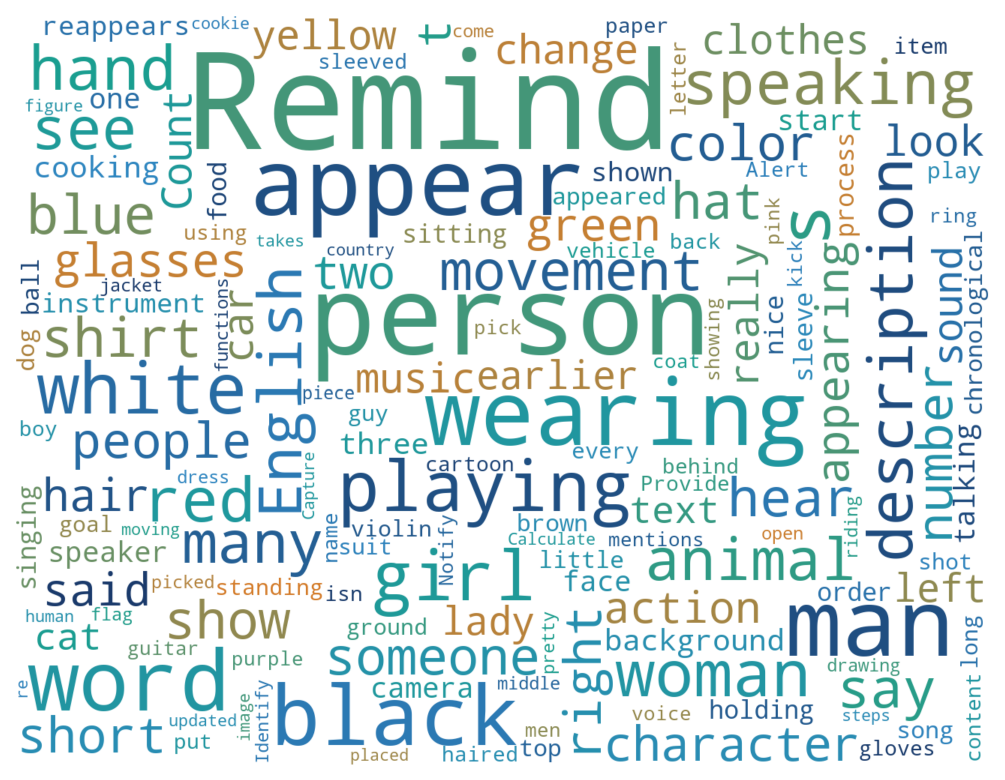}
    \end{subfigure}

    \caption{
    Overview of the dataset characteristics:
    (a) Distribution of video durations;
    (b) Distribution of video categories;
    (c) Linguistic characteristics of text queries.
    }

    \label{fig:data-overview}
\end{figure}

\name{} consists of 660 videos paired with human-curated question–answer annotations, spanning diverse domains such as education, entertainment, sports, and daily activities (Figure~\ref{fig:data-overview}(b)). All videos are under one minute in length, with an average duration of 34 seconds; the distribution of video durations is shown in Figure~\ref{fig:data-overview}(a). All questions are open-ended to better reflect real-world usage. The linguistic characteristics of the queries are illustrated in Figure~\ref{fig:data-overview}(c).

\subsection{Evaluation Pipeline}

Existing evaluations focus mainly on answer correctness, overlooking \textit{when} a response is produced. In \name, we introduce an LLM-as-a-Judge framework that jointly evaluates response timing and content correctness. Since Real-Time Description (RTD) and Proactive Reminder (PR) follow different response patterns, we design separate evaluation strategies for the two scenarios. In the following, we briefly describe the evaluation pipeline for each scenario.

\begin{figure}[htbp]       
    \centering            
    \includegraphics[width=\textwidth]{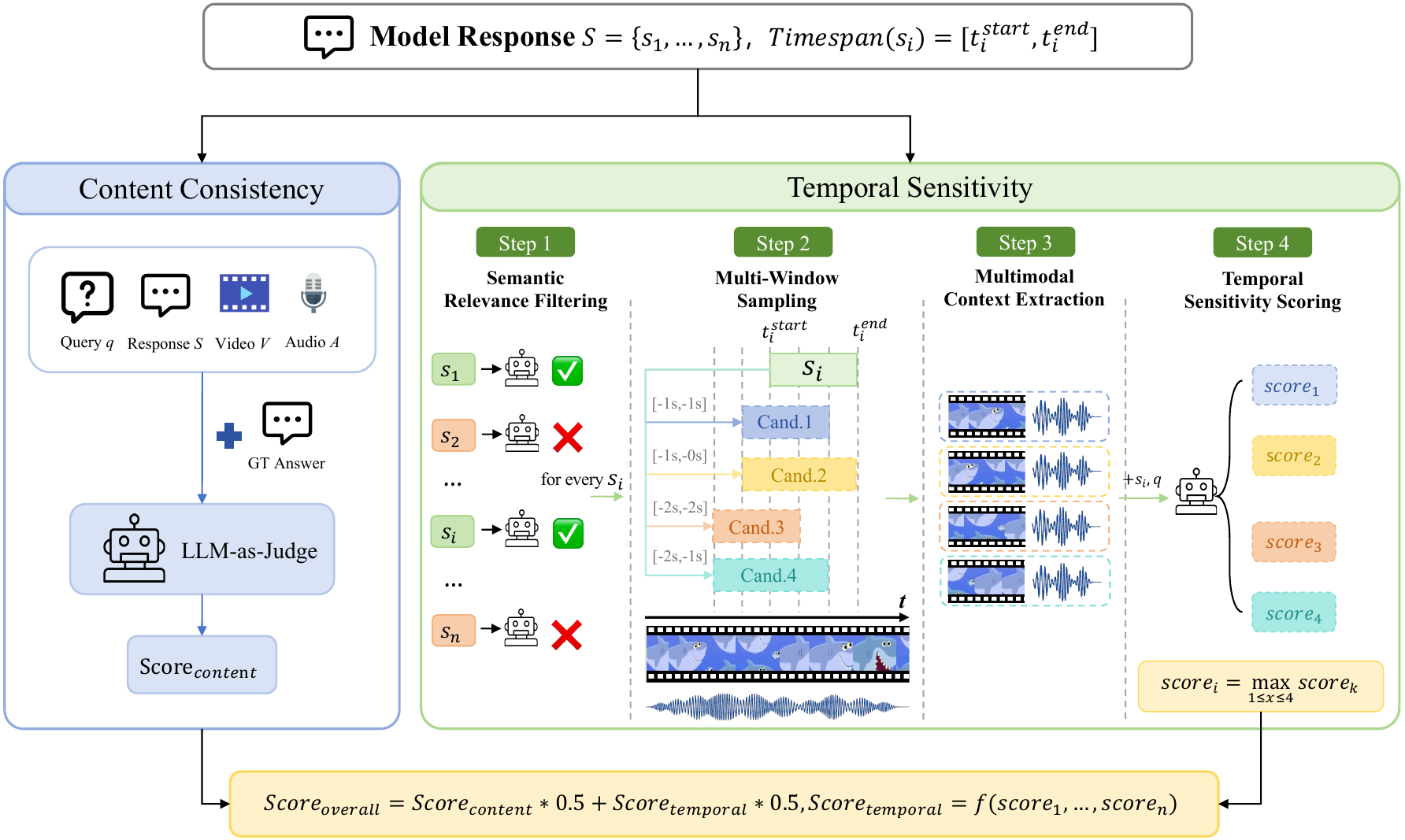}
    \caption{\textbf{The automatic evaluation pipeline for Real-Time Description.} The framework assesses two dimensions: Content Consistency for global quality, and Temporal Sensitivity for streaming alignment. The final score is computed as a weighted combination of the two.}  
    \label{fig:real-time-pipeline}   
\end{figure}

\subsubsection{Real-Time Description}

Real-Time Description requires models to generate continuous, streaming descriptions synchronized with evolving video content. This scenario evaluates temporal alignment at sentence-level granularity. To this end, we adopt a two-dimensional evaluation framework consisting of Content Consistency and Temporal Sensitivity. Given a user query $q$ and a model’s streaming output $S = \{s_1, s_2, \ldots, s_n\}$, each sentence $s_i$ is associated with a time interval $[t_i^{\text{start}},\, t_i^{\text{end}}]$, enabling fine-grained evaluation along both dimensions. The evaluation pipeline is illustrated in Figure~\ref{fig:real-time-pipeline}.

\paragraph{Content Consistency}

This metric focuses on global semantic alignment between the model response and the omni-modal input. We extract the full video and corresponding audio, and employ an LLM-as-a-Judge framework to assess whether the response is consistent with the user query and the underlying video–audio content, yielding the content consistency score, $Score_{\text{content}}$. The evaluation follows a score-deduction scheme, penalizing factual errors, hallucinations, and omissions.



\paragraph{Temporal Sensitivity}

Temporal Sensitivity measures whether the model captures real-time changes and generates timely, instruction-aligned responses. However, raw streaming outputs contain two sources of noise: (1) irrelevant utterances (e.g., polite phrases) that should not be temporally evaluated, and (2) natural latency variations in model response timing. To address these, we introduce a four-step evaluation pipeline.

\textit{Semantic Relevance Filtering:} To exclude non-substantive outputs from temporal assessment, each sentence \(s_i\) is classified as relevant or irrelevant by an LLM-as-a-Judge framework based on user instruction and video–audio context. Let \(S_{\text{irr}} \subseteq S\) denote irrelevant sentences. These are excluded from evaluation, and their proportion \(r = |S_{\text{irr}}| / |S|\) attenuates the final score.

\textit{Multi-Window Sampling:} To tolerate natural perception-to-generation latency (empirically \(\approx 2\) seconds) while penalizing clearly mistimed responses, we construct $k = 4$ candidate windows around each original timespan \([t_i^{\text{start}}, t_i^{\text{end}}]\). They are
\(w_1: [t_i^{\text{start}}-1,\ t_i^{\text{end}}-1],\ w_2: [t_i^{\text{start}}-2,\ t_i^{\text{end}}-1],\ w_3: [t_i^{\text{start}}-2,\ t_i^{\text{end}}-2],\ w_4: [t_i^{\text{start}}-1,\ t_i^{\text{end}}]\).

\textit{Multimodal Context Extraction \& Scoring:} For each candidate window \(w\), we sample video frames at 2 FPS and extract the corresponding audio segment. An LLM judge then evaluates alignment between sentence \(s_i\) and each window. The sentence score is the maximum alignment score across these windows.
\begin{equation}
\mathrm{score}(s_i) = \max_{k \in \{1,2,3,4\}} \mathrm{LLM}(q, s_i, video_{w_k}, audio_{w_k})
\label{eq:t1}
\end{equation}

The final Temporal Sensitivity score averages over relevant sentences with an attenuation penalty:
\begin{equation}
\mathrm{Score}_{\text{temporal}} = \left( \frac{1}{|S_{\text{rel}}|} \sum_{s_i \in S_{\text{rel}}} \mathrm{score}(s_i) \right) \times (1 - \lambda \cdot r)
\end{equation}
where \(S_{\text{rel}} = S \setminus S_{\text{irr}}\). \(\lambda\) is a hyperparameter controlling the penalty intensity and we set $\lambda = 1$. The overall score combines Content Consistency and Temporal Sensitivity equally:
\begin{equation}
\mathrm{Score}_{\text{overall}} = 0.5 \cdot \mathrm{Score}_{\text{content}} + 0.5 \cdot \mathrm{Score}_{\text{temporal}}
\end{equation}
Each metric is reported on a 0 – 3 scale, then linearly mapped to 0 – 100.

To improve alignment with human judgments, we experimented with multiple iterative design strategies for our evaluation framework. 
Overall, our evaluation framework shows strong agreement with human judgments.
Detailed ablation and analysis of these iterations, including comparisons with human annotations, are provided in Appendix~\ref{sec:iterations}.

\subsubsection{Proactivate Reminder}



Proactive Reminder evaluates the ability to identify relevant events and determine appropriate response timing under streaming video inputs. \name{} provides annotated timestamps for each event. During evaluation, we extract the model’s responses within a fixed 10-second window following each event timestamp and assess them using an LLM-as-a-Judge framework. The evaluation focuses on both event identification and the consistency of the response with the user instruction. For Correction tasks, the evaluation measures whether the model accurately revises the user’s description based on the video content. For Event Reminder and Post-Event Reminder tasks, it assesses whether the model produces appropriate responses when the event occurs. In addition, for samples where the reminder event occurs multiple times, the model must correctly respond to all occurrences for the sample to be considered successful. In practice, we employ Gemini-3-Flash-thinking as the LLM judge. Implementation details, including the prompts, are provided in Appendix~\ref{sec:eval_details}.

\section{Experiments}

\subsection{Baselines}
\label{sec:exp_config}

We focus on evaluating multimodal models that support duplex inference. Specifically, we include LiveCC (Base/Instruct)~\citep{chen2025livecc}, MMDuet2~\citep{wang2025mmduet2}, StreamingVLM~\citep{xu2025streamingvlm}, and MiniCPM-o 4.5~\citep{yao2024minicpm}. 
All experiments are conducted on a single NVIDIA A100 GPU. 
For each model, we follow its native duplex inference protocol to obtain real-time responses. Outputs are recorded as they are emitted over time, enabling evaluation of response timing and interaction behavior under streaming conditions.

\paragraph{Human Evaluation}



We conduct two human tests under different protocols. \textbf{Human-Duplex}. We sample 20 instances per scenario, covering all sub-tasks. Four independent annotators not involved in dataset construction, provide real-time spoken responses while watching each video for the first time. Responses are recorded with start times strictly synchronized to video playback, following the same streaming protocol as model inference. This evaluation reflects human performance under real-time constraints. \textbf{Human-Offline}. To assess the upper bound of content understanding without temporal pressure, we conduct an offline human study. Annotators are allowed to preview the entire video and instruction beforehand, and then generate a complete response without real-time streaming constraints—mirroring the inference paradigm of offline MLLMs. This provides a reference for evaluating the content accuracy ceiling when timing is not a factor.

\begin{table}[htbp]
\caption{Performance of duplex models on \name{}. 
We report per-task scores, scenario-level averages, and the overall benchmark score. The best-performing results are shown in bold.}
\label{tab:main-results}
\centering
\renewcommand{\arraystretch}{1.4}
\resizebox{\textwidth}{!}{
\begin{tabular}{l c c c c c c c c c c c c}
\toprule
\multirow{2}{*}{\textbf{Models}} & \multicolumn{7}{c}{\textbf{Real-Time Description}} & \multicolumn{4}{c}{\textbf{Proactive Reminder}} & \multirow{2}{*}{\textbf{Avg.}}\\
\cmidrule(lr){2-8}
\cmidrule(lr){9-12}
& CT & IR & Omni & WK & OCR & FM & Avg. & ER & PER & CR & Avg.\\
\midrule
\rowcolor{gray!20}
Human-Offline & 82.6 & 74.6 & 88.9 & 83.8 & 88.8 & 79.5 & 83.0 & 100.0 & 100.0 & 100.0 & 100.0 & 91.5 \\ 
\rowcolor{gray!20}
Human-Duplex &68.8 & 81.1 & 64.1 & 73.2 & 65.1 & 72.2 & 70.8 & 89.3 & 97.9 & 91.1 & 92.8 & 81.8 \\ 
\midrule
LiveCC-Base~\citep{chen2025livecc} & 17.6 & 36.7 & 32.2 & 39.3 & 49.5 & 33.4 & 34.8 & 3.1 & 1.0 & 1.5 & 1.9 & 18.4 \\
StreamingVLM~\citep{xu2025streamingvlm} & 29.2 & 34.0 & 36.7 & 39.4 & 42.9 & 35.1 & 36.2 & 1.6 & 0.0 & 3.0 & 1.7 & 19.0\\
LiveCC-Inst~\citep{chen2025livecc} & 30.9 & 35.3 & 48.6 & 47.7 & 52.7 & 41.9 & 42.9 & 7.8 & 2.0 & 3.8 & 4.7 & 23.8 \\
MMDuet2~\citep{wang2025mmduet2} & \textbf{53.3} & 54.4 & 57.6 & 59.6 & 64.9 & \textbf{60.8} & 58.4 & \textbf{24.2} &9.1 &2.3 &11.9 &35.2 \\
MiniCPM-o 4.5~\citep{yao2024minicpm} & 51.4 & \textbf{58.2} & \textbf{58.4} & \textbf{63.7} & \textbf{68.6} & 54.3 & \textbf{59.1} & 18.8 & \textbf{11.1} & \textbf{27.8} & \textbf{20.0} & \textbf{39.6}\\
\bottomrule
\end{tabular}
}
\end{table}

\subsection{Main results}
Table~\ref{tab:main-results} summarizes the performance of duplex models on the Real-Time Description and Proactive Reminder scenarios. Our primary findings are as follows:


\textbf{Significant Gap Between Models and Human Performance.}
Overall, current duplex models fall substantially short of human performance on \name{}, with the best model achieving 39.6 compared to 81.8 for Human-Duplex. While MiniCPM-o 4.5 consistently outperforms other models, all systems remain far from human-level real-time interaction. Across models, performance is noticeably higher on Real-Time Description than on Proactive Reminder, indicating a shared difficulty in handling event-driven interaction. This suggests that, although models can partially track evolving content, they struggle more fundamentally with deciding when to respond.

\begin{table}[htbp]
\caption{Performance of duplex models on the six sub-tasks of Real-Time Description, including scores for two evaluation dimensions and an overall aggregated score.}
\label{tab:real-time description}
\centering
\small
\renewcommand{\arraystretch}{1.1}
\resizebox{\textwidth}{!}{
\begin{tabular}{l l c c c c c c c}
\toprule
\multirow{2}{*}{Models} & \multirow{2}{*}{Metric} & \multicolumn{7}{c}{Task} \\
\cmidrule(lr){3-9}
& & CT & IR & Omni & WK & OCR & FM & Avg. \\
\midrule
\multirow{3}{*}{Human-Offline}
& Temporal Sensitivity  & 73.5 & 87.4 & 83.3 & 82.2 & 91.5 & 87.8 & 84.3 \\
& Content Consistency  & 91.7 & 61.7 & 94.4 & 85.4 & 86.1 & 71.1 & 81.7 \\
& Average & 82.6 & 74.6 & 88.9 & 83.8 & 88.8 & 79.5 & 83.0 \\ 
\hline
\multirow{3}{*}{Human-Duplex}
& Temporal Sensitivity  & 77.9 & 91.7 & 72.5 & 83.5 & 68.8 & 85.7 & 80.0 \\
& Content Consistency  & 59.7 & 70.5 & 55.7 & 62.9 & 61.4 & 58.7 & 61.5 \\
& Average & 68.8 & 81.1 & 64.1 & 73.2 & 65.1 & 72.2 & 70.8 \\ 
\hline
\multirow{3}{*}{LiveCC-Base~\citep{chen2025livecc}}
& Temporal Sensitivity  & \cellcolor{blue!10}25.8 & \cellcolor{blue!10}56.9 & \cellcolor{blue!10}45.9 & \cellcolor{blue!10}56.6 & \cellcolor{blue!10}63.1 & \cellcolor{blue!10}47.5 & \cellcolor{blue!10}49.3 \\
& Content Consistency  & \cellcolor{red!10}9.4 & \cellcolor{red!10}16.5 & \cellcolor{red!10}18.5 & \cellcolor{red!10}21.9 & \cellcolor{red!10}35.8 & \cellcolor{red!10}19.3 & \cellcolor{red!10}20.2 \\
& Average & 17.6 & 36.7 & 32.2 & 39.3 & 49.5 & 33.4 & 34.8 \\
\hline
\multirow{3}{*}{StreamingVLM~\citep{xu2025streamingvlm}}
& Temporal Sensitivity  & \cellcolor{blue!10}47.0 & \cellcolor{blue!10}57.9 & \cellcolor{blue!10}58.8 & \cellcolor{blue!10}55.6 & \cellcolor{blue!10}55.9 & \cellcolor{blue!10}54.8 & \cellcolor{blue!10}55.0 \\
& Content Consistency  & \cellcolor{red!10}11.3 & \cellcolor{red!10}10.1 & \cellcolor{red!10}14.5 & \cellcolor{red!10}23.2 & \cellcolor{red!10}29.8 & \cellcolor{red!10}15.4 & \cellcolor{red!10}17.4 \\
& Average & 29.2 & 34.0 & 36.7 & 39.4 & 42.9 & 35.1 & 36.2 \\
\hline
\multirow{3}{*}{LiveCC-Inst~\citep{chen2025livecc}}
& Temporal Sensitivity  & \cellcolor{blue!10}49.9 & \cellcolor{blue!10}57.8 & \cellcolor{blue!10}75.5 & \cellcolor{blue!10}63.5 & \cellcolor{blue!10}64.8 & \cellcolor{blue!10}63.2 & \cellcolor{blue!10}62.5 \\
& Content Consistency  & \cellcolor{red!10}11.9 & \cellcolor{red!10}12.8 & \cellcolor{red!10}21.7 & \cellcolor{red!10}31.8 & \cellcolor{red!10}40.5 & \cellcolor{red!10}20.5 & \cellcolor{red!10}23.2 \\
& Average & 30.9 & 35.3 & 48.6 & 47.7 & 52.7 & 41.9 & 42.9 \\
\hline

\multirow{3}{*}{MMDuet2~\citep{wang2025mmduet2}}
& Temporal Sensitivity  & \cellcolor{blue!10}77.8 & \cellcolor{blue!10}82.7 & \cellcolor{blue!10}82.7 & \cellcolor{blue!10}73.6 & \cellcolor{blue!10}78.4 & \cellcolor{blue!10}80.2 & \cellcolor{blue!10}79.2 \\
& Content Consistency  & \cellcolor{red!10}28.8 & \cellcolor{red!10}26.0 & \cellcolor{red!10}32.4 & \cellcolor{red!10}45.6 & \cellcolor{red!10}51.4 & \cellcolor{red!10}41.4 & \cellcolor{red!10}37.6 \\
& Average & 53.3 & 54.4 & 57.6 & 59.6 & 64.9 & 60.8 & 58.4 \\
\hline
\multirow{3}{*}{MiniCPM-o 4.5~\citep{yao2024minicpm}}
& Temporal Sensitivity  & \cellcolor{blue!10}70.1 & \cellcolor{blue!10}84.3 & \cellcolor{blue!10}81.3 & \cellcolor{blue!10}82.5 & \cellcolor{blue!10}84.6 & \cellcolor{blue!10}76.3 & \cellcolor{blue!10}79.9 \\
& Content Consistency  & \cellcolor{red!10}32.7 & \cellcolor{red!10}32.0 & \cellcolor{red!10}35.5 & \cellcolor{red!10}44.9 & \cellcolor{red!10}52.6 & \cellcolor{red!10}32.3 & \cellcolor{red!10}38.3 \\
& Average & 51.4 & 58.2 & 58.4 & 63.7 & 68.6 & 54.3 & 59.1 \\
\bottomrule
\end{tabular}
}
\end{table}

\begin{wrapfigure}{r}{0.4\linewidth}
    \centering
    \includegraphics[width=\linewidth]{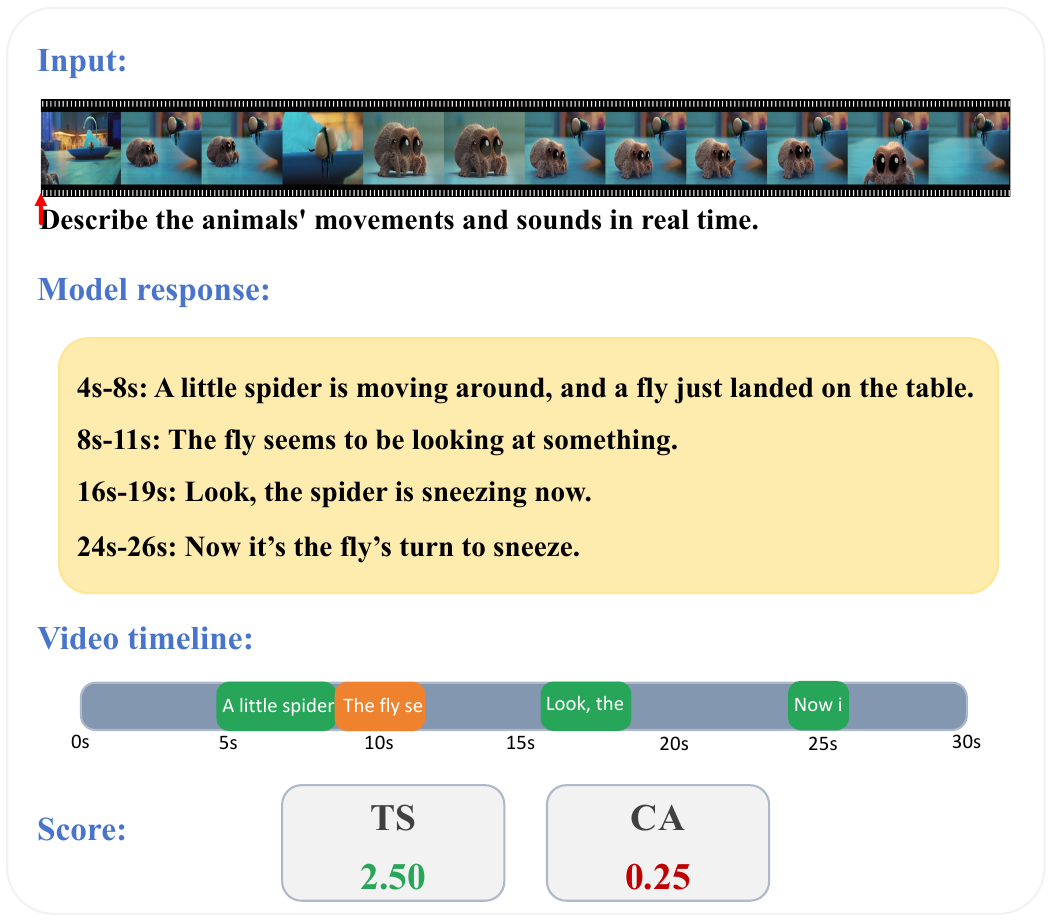}
    \caption{Example of model predictions in Real-Time Description.}
    \label{fig:case_study}
\end{wrapfigure}

\textbf{Models Excel at Perception but Struggle with Structured Reasoning.} Fine-grained analysis reveals a clear gap between perception and reasoning abilities. While models perform relatively well on low-level tasks such as OCR and fine-grained motion (e.g., MiniCPM-o 4.5 achieves 68.6 on OCR), performance drops on tasks requiring structured reasoning. In particular, Counting is consistently the most challenging task across models (e.g., 51.4 for MiniCPM-o 4.5), with lower scores also observed on Interaction Relationships and World Knowledge. These results suggest that current duplex models remain limited in integrating dynamic context into coherent reasoning.

\textbf{Models Produce Sparse Responses, Limiting Holistic Understanding.} 
Our analysis reveals a clear discrepancy between local and global evaluation dimensions. As shown in Table~\ref{tab:real-time description}, models achieve relatively strong performance in Temporal Sensitivity but consistently underperform in Content Consistency. 
This gap mainly stems from the output behavior of current models: they tend to generate sparse and intermittent responses, remaining silent for a large portion of the video and producing outputs only occasionally. While such behavior may help maintain local temporal alignment, it often fails to capture the continuous context of the video, leading to poor global consistency.
Figure~\ref{fig:case_study} further illustrates this pattern, where outputs are temporally sparse and fragmented. These results suggest that current models struggle to reconcile timely response generation with holistic content understanding, highlighting a fundamental limitation in real-time duplex interaction.
\begin{wraptable}{r}{0.6\linewidth}
\vspace{-10pt}
\centering
\caption{Distribution of error types for model responses under the Proactive Reminder setting (in percentage, \%).}
\label{tab:eval-stats}
\small
\setlength{\tabcolsep}{3pt}
\begin{tabular}{lccc}
\toprule
\textbf{Models} & \textbf{No Answer} & \textbf{Partially Correct} & \textbf{Wrong} \\
\midrule
LiveCC-Base~\citep{chen2025livecc}   & 5.8 & 1.2 & 91.1 \\
StreamingVLM~\citep{xu2025streamingvlm}  & 0.8 & 0.8 & 96.7 \\
LiveCC-Inst~\citep{chen2025livecc}   & 0.8 & 1.4 & 93.1 \\
MMDuet2~\citep{wang2025mmduet2}       & 75.8 & 5.1 & 7.2  \\
MiniCPM-o 4.5~\citep{yao2024minicpm} & 49.2 & 3.6 & 27.2 \\
\bottomrule
\end{tabular}
\vspace{-10pt}
\end{wraptable}
\textbf{Models Fail to Determine When to Respond in Proactive Reminder.}
From the results in Table~\ref{tab:main-results}, we observe that the best-performing model achieves only 20.0, indicating that the overall performance remains very limited. We further analyze the error distribution, as shown in Table~\ref{tab:eval-stats}. MiniCPM-o 4.5 and MMDuet2 are dominated by No Answer cases. In contrast, LiveCC and StreamingVLM mainly produce Wrong outputs. 
We further analyze the underlying causes and find that these models often generate continuous caption-like descriptions without following the instruction or identifying relevant events. This suggests that they fail to determine when a response should be triggered.
Moreover, even when models correctly detect events, maintaining content consistency remains challenging.
Overall, these results point to a fundamental limitation of current duplex MLLMs: the inability to decide when to respond. 

\section{Conclusion and Future Work}

We introduce \name{}, the first benchmark for evaluating real-time full-duplex capabilities of omni-modal models. The benchmark comprises two scenarios: Real-Time Description (six tasks) and Proactive Reminder (three tasks), with 660 videos and human-curated timestamp-level annotations. 
Our experiments reveal two key findings. For Real-Time Description, models fail to balance global content consistency with local temporal sensitivity. For Proactive Reminder, models struggle to determine when to respond. 
These results further highlight the importance of real-time duplex interaction capabilities.
We hope this work will facilitate future research toward more capable real-time duplex multimodal systems.

Future work may extend \name{} toward longer and more complex interaction settings. As duplex multimodal systems continue to evolve, we also expect future benchmarks to cover richer modalities and broader forms of real-time interaction.

\medskip





\bibliography{neurips_2026}
\bibliographystyle{unsrt}
\newpage
\appendix

\section{Detailed Evaluation Protocols}
\label{sec:eval_details}

This section provides the complete evaluation protocols for the Real-Time Description and Proactive Reminder.

\subsection{Content Consistency}

Content Consistency measures the factual consistency between the model-generated response and the video content, while ensuring alignment with user instructions.

\subsubsection{Evaluation Process}

The evaluation follows a deduction-based scoring mechanism:

\begin{enumerate}
    \item The evaluator starts from a perfect score of \textbf{3.00}.
    \item For each error identified, a specific penalty is deducted according to Table~\ref{tab:content_penalties}.
    \item The final score is the maximum of the calculated result and \textbf{0.01}, unless the response is completely empty or entirely irrelevant, in which case the score is \textbf{0.00}.
\end{enumerate}

\subsubsection{Penalty Table}

\begin{table}[h]
\caption{Content Consistency Penalty Values}
\label{tab:content_penalties}
\centering
\begin{tabular}{l l c}
\toprule
\textbf{Error Category} & \textbf{Severity} & \textbf{Penalty} \\
\midrule
Critical Factual Error (wrong object/action/color/count) & High & -1.00 \\
Critical Factual Error (partially wrong, e.g., "dark blue" vs "navy blue") & Medium & -0.75 \\
Minor Factual Error & Low & -0.25 \\
Hallucination (describing non-existent content) & Severe & -1.50 \\
Key Information Omission (missing main step/element) & High & -0.75 \\
Minor Detail Omission & Low & -0.25 \\
Vagueness ("mixes something" vs specific action) & Medium & -0.50 \\
Repetition & Low & -0.10 \\
Irrelevant Content & Medium & -0.50 \\
\bottomrule
\end{tabular}
\end{table}

\subsubsection{Evaluation Prompt}

The following prompt is used for Content Consistency evaluation:

\begin{tcolorbox}[colback=gray!5, colframe=black, title=Content Consistency Evaluation Prompt]
You are a PRECISE CONTENT ACCURACY EVALUATOR. Your task is to assign a 
FINE-GRAINED decimal score (0.00-3.00) based on the model's response accuracy.\\

Basic Information:\\
- Original user instruction: {question}\\
- Model generated response: "{response}"\\
- Reference annotations: {gt\_texts} (if provided)\\

Scoring Rules:\\
1. Start from 3.00\\
2. Deduct penalties for each error\\
3. Output ONLY JSON with "content\_score" and "content\_reasoning"
\end{tcolorbox}

\subsection{Temporal Sensitivity}

Temporal Sensitivity measures the alignment between the model-generated text and the video's temporal windows—specifically, whether the model describes the corresponding video content at the appropriate time.

\subsubsection{Evaluation Process}

The metric evaluates a timestamped response format \(S = \{s_1, s_2, \ldots, s_n\}\) with each sentence \(s_i\) associated with a time interval $[t_i^{\text{start}},\, t_i^{\text{end}}]$.The Temporal Sensitivity evaluation consists of four steps:

\paragraph{Step 1: Semantic Relevance Filtering}
Each sentence \(s_i\) with timestamp \((\text{start}_i, \text{end}_i)\) is classified as \textbf{relevant} or \textbf{irrelevant}. Irrelevant sentences (e.g., polite phrases like "No problem," "I'm happy to help") are excluded from temporal evaluation. The proportion of irrelevant sentences \(r = |S_{\text{irr}}| / |S|\) is used for score attenuation.

\paragraph{Step 2: Multi-Window Sampling}
Based on the empirical observation that a 2-second perception-to-generation latency (\(d = 2\)) is reasonable for streaming models, four candidate windows are constructed for each relevant sentence:
\(w_1: [t_i^{\text{start}}-1,\ t_i^{\text{end}}-1],\ w_2: [t_i^{\text{start}}-2,\ t_i^{\text{end}}-1],\ w_3: [t_i^{\text{start}}-2,\ t_i^{\text{end}}-2],\ w_4: [t_i^{\text{start}}-1,\ t_i^{\text{end}}]\).

These windows account for potential latency variations around the assumed optimal delay.

\paragraph{Step 3: Multimodal Context Extraction}
For each candidate window \(w_k (k \in [1,4])\), the corresponding audio segment is extracted, and video frames are sampled at \(f = 2\) frames per second.

\paragraph{Step 4: Scoring}
An LLM judge evaluates the alignment between the sentence content and each candidate window, considering both visual and audio modalities. The sentence score $score(s_i)$ is the maximum across all windows, as shown in Equation~\ref{eq:t1}.


\subsubsection{Final Score Calculation}

The final Temporal Sensitivity score is computed as:

\begin{equation}
\mu_{\text{rel}} = \frac{1}{|S_{\text{rel}}|} \sum_{s \in S_{\text{rel}}} \text{score}(s)
\end{equation}

\begin{equation}
S_{\text{temporal}} = \mu_{\text{rel}} \times (1 - \lambda \cdot r)
\end{equation}

where:
\begin{itemize}
    \item \(\mu_{\text{rel}}\) is the average score of relevant sentences (0-3 scale);
    \item \(r\) is the proportion of irrelevant sentences;
    \item \(\lambda\) is a hyperparameter controlling the penalty intensity. We set $\lambda = 1$.
\end{itemize}

\subsubsection{Temporal Sensitivity Scoring Guidelines}

\begin{table}[h]
\centering
\caption{Temporal Sensitivity Scoring Criteria}
\label{tab:temporal_criteria}
\begin{tabular}{c p{10cm}}
\toprule
\textbf{Score} & \textbf{Definition} \\
\midrule
3 & Excellent temporal alignment. Response accurately describes the current video segment and aligns perfectly with the instruction. \\
2 & Moderate temporal alignment. Response is generally accurate but has minor inaccuracies or contains some descriptions of other time periods. \\
1 & Poor temporal alignment. Response has significant issues, largely describes wrong time periods, or is mostly irrelevant. \\
0 & No temporal alignment. Response is completely irrelevant or describes completely wrong time periods. \\
\bottomrule
\end{tabular}
\end{table}

\subsubsection{Relevance Classification Guidelines}

\begin{table}[h]
\centering
\caption{Relevance Classification Criteria}
\label{tab:relevance_criteria}
\begin{tabular}{c p{10cm}}
\toprule
\textbf{Label} & \textbf{Definition} \\
\midrule
Relevant (1) & Contains substantive content responding to the instruction or describing video content. \\
Irrelevant (0) & Polite phrases, acknowledgments, thinking pauses, or generic responses without substantive content. \\
\bottomrule
\end{tabular}
\end{table}

\subsubsection{Evaluation Prompt for Temporal Sensitivity}

\begin{tcolorbox}[colback=gray!5, colframe=black, title=Temporal Sensitivity Evaluation Prompt]
You are a professional evaluator for real-time multimodal systems.

Basic Information:\\
- Analysis Time Range: Video segment from {start}s to {end}s\\
- Original Instruction: {question}\\
- Response to Evaluate: "{sentence}"\\

Scoring Guidelines:\\
- 3: Excellent temporal alignment\\
- 2: Moderate temporal alignment\\
- 1: Poor temporal alignment\\
- 0: No temporal alignment\\

Answer Relevance Classification:\\
- Relevant (1): Contains substantive content\\
- Irrelevant (0): Only polite phrases or acknowledgments\\

Output JSON:\\
\{\\
    "temporal\_score": <0-3>,\\
    "temporal\_reasoning": "<explanation>",\\
    "is\_relevant": <0 or 1>\\
\}
\end{tcolorbox}

\subsection{Proactive Reminder Evaluation}

Proactive Reminder evaluates the ability to identify relevant events and determine appropriate response timing. The evaluation follows a two-stage pipeline: (1) temporal window extraction and (2) LLM-based judgment.

\subsubsection{Temporal Window Extraction}

Given annotated event timestamps from the ground truth, we extract the model's response within a fixed time window following each event occurrence. Specifically, for each reminder event with start time \(t_{event}\), we collect all model-generated sentences whose timestamp falls within \([t_{event}, t_{event} + \Delta]\), where \(\Delta = 10\) seconds is the evaluation window. These collected sentences are concatenated to form the response segment for that event.

\begin{equation}
\text{response\_segment}(t_{event}) = \bigcup \{ s_i \mid t_{event} \le \text{time}(s_i) \le t_{event} + \Delta \}
\end{equation}

\subsubsection{LLM Judgment Prompt}

The extracted response segment is evaluated by an LLM-as-a-Judge framework. The prompt varies by task type.

\textbf{Event Reminder \& Post-Event Reminder Prompt:}

\begin{tcolorbox}[colback=gray!5, colframe=black, title=Event Reminder Evaluation Prompt]
Please determine whether the model's output correctly completes the reminder task.

\textbf{Task Description:}\\
The user gives an instruction to remind them when a specific event occurs.\\
We have extracted the model's text generated immediately after the event (within a 10-second window).\\
Determine whether the text clearly demonstrates that "the event has occurred" as a reminder.\\

\textbf{Input:}\\
- User instruction: \{instruction\}\\
- Model output: "\{response\}"\\

\textbf{Criteria:}\\
The model output must satisfy:\\
1. Clearly refer to the target event described in the instruction;\\
2. Express an intention to remind or inform that the event has occurred;\\
3. Not be vague or unrelated to the event;\\
4. If the output is ambiguous, misidentifies the event, or does not mention the event, it is considered a failure.\\

\textbf{Scoring:}\\
- 1 = Successful reminder (explicitly mentions the event and completes the reminder)\\
- 0 = Unsuccessful reminder (vague / incorrect / event not mentioned)\\

\textbf{Output Format:}\\
Only output JSON:\\
\{\\
    "success\_score": <0 or 1>,\\
    "reasoning": "<explanation>"\\
\}
\end{tcolorbox}

\textbf{Correction Task Prompt:}

\begin{tcolorbox}[colback=gray!5, colframe=black, title=Correction Task Evaluation Prompt]
Please determine whether the model's output correctly completes the correction task.

\textbf{Task Description:}\\
The user provides an instruction or statement that contains incorrect information.\\
The system needs to identify the error and provide the correct information.\\

\textbf{Input:}\\
- User instruction: \{instruction\}\\
- Ground truth answer: \{ground\_answer\}\\
- Model output: "\{response\}"\\

\textbf{Criteria:}\\
1. Compare the user instruction with the ground truth answer to identify the error(s).\\
2. Check whether the model output corrects these error(s) consistent with the ground truth.\\
3. The correction must maintain correct context (e.g., subject, object) consistent with both instruction and answer.\\
4. Extra information unrelated to correction should be ignored, unless it contradicts the instruction or answer.\\

\textbf{Scoring:}\\
- 1 = Successful correction (all errors corrected with consistent context)\\
- 0 = Unsuccessful correction (missing errors, inconsistent correction, or context mismatch)\\

\textbf{Output Format:}\\
Only output JSON:\\
\{\\
    "success\_score": <0 or 1>,\\
    "reasoning": "<explanation>"\\
\}
\end{tcolorbox}

\subsubsection{Final Score Calculation}

For a sample containing \(N\) events (reminders), let \(score_j \in \{0,1\}\) be the LLM judgment for event \(j\). The sample-level total score is defined as:

\begin{equation}
\text{Score}_{\text{sample}} = \mathbf{1}\left[ \sum_{j=1}^{N} score_j = N \right]
\end{equation}

where \(\mathbf{1}[\cdot]\) is the indicator function. That is, the sample is considered successful \textit{only if all events are correctly handled}. This strict criterion reflects the real-world requirement for reliable proactive systems.

The overall model performance on a task is the average of sample-level scores across all samples in that task.

\section{Iterative Design and Human Alignment Analysis}
\label{sec:iterations}

\subsection{Motivation}

\name{} evaluates open-ended responses without objective ground-truth answers. To ensure that our automatic evaluation framework aligns with human perception, we constructed a calibration set with the help of human annotators and iteratively refined our evaluation prompts and strategies. The Spearman correlation between automatic evaluation scores and human judgments serves as the alignment metric throughout this process.

\subsection{Calibration Set Construction}

To systematically calibrate the two evaluation metrics (Content Consistency and Temporal Sensitivity), we constructed a calibration set following a controlled design. For a given video-question instance, we generated responses that vary the scores of the two metrics in a structured manner.

Specifically, we fixed two metrics at their maximum score (3.00) while varying the remaining metric across the full range (0, 1, 2, 3). This yielded the following 7 distinct score combinations (ordered as \textit{Temporal Sensitivity} - \textit{Content Consistency} :

In addition, two reference ground-truth responses were included as baselines. In total, the calibration set comprises 7 distinct video-question instances, yielding 63 annotated answer samples (7 instances × 9 responses per instance). Each response was manually annotated by human evaluators to establish reference scores for all three metrics.

\subsection{Iterative Refinement and Results}

\subsubsection{Content Consistency}

For Content Consistency, we experimented with:
\begin{itemize}
    \item Different frame sampling rates (0.5 FPS vs. 0.3333 FPS)
    \item Different numbers of ground-truth references (0, 1, or 2 GT files)
    \item Prompt refinements to improve scoring precision
\end{itemize}

\begin{table}[h]
\centering
\caption{Content Consistency Iteration Results}
\label{tab:content_iterations}
\begin{tabular}{l c}
\toprule
\textbf{Configuration} & \textbf{Spearman Correlation} \\
\midrule
0 GT, 0.5 FPS & 0.8258 \\
1 GT, 0.5 FPS & 0.8368 \\
2 GT, 0.5 FPS & 0.8969 \\
0 GT, 0.3333 FPS & 0.8164 \\
1 GT, 0.3333 FPS & 0.8573 \\
2 GT, 0.3333 FPS & \textbf{0.9165} \\
\bottomrule
\end{tabular}
\end{table}

The best alignment was achieved with \textbf{2 GT references at 0.3333 FPS}.





\subsubsection{Temporal Sensitivity}

For Temporal Sensitivity, we explored multiple strategies:

\begin{itemize}
    \item \textbf{Window Strategy}: Compared single-window (shifting start/end by -2 seconds) against four-window sampling (shifting start/end by -1/-2 seconds) to tolerate reasonable perception-to-generation latency.
    \item \textbf{Unit of Analysis}: Compared sentence-level segmentation (sentence-ctc) against action-level segmentation (action-ctc) based on semantic boundaries.
    \item \textbf{Modality for Alignment}: Compared using video frames (2 FPS) as context versus using ground-truth text with character-level timestamps as an oracle reference.
    \item \textbf{Prompt Refinement}: Iteratively adjusted LLM judge prompts based on disagreement analysis from human re-annotation.
    \item \textbf{Irrelevant Sentence Penalty}: Introduced attenuation factor \(\lambda=1\) to penalize polite phrases and non-substantive responses.
    \item \textbf{Sampling Rate}: Compared video frame sampling at 2 FPS versus 3 FPS for context extraction.
    \item \textbf{Window Selection Policy}: Adjusted candidate window offsets from symmetric shifts to an optimized asymmetric strategy favoring slightly delayed responses.
\end{itemize}

\begin{table}[h]
\centering
\caption{Temporal Sensitivity Iteration Results}
\label{tab:temporal_iterations}
\begin{tabular}{l c c}
\toprule
\textbf{Configuration} & \textbf{Variant A} & \textbf{Variant B} \\
\midrule
Window Strategy & Single-window & $\rightarrow$ Four-window \\
& (0.7021) & (0.7343) \\
\midrule
Unit of Analysis & Action-level & $\rightarrow$ Sentence-level \\
& (0.6841 / 0.5781) & (0.7343) \\
\midrule
Modality for Alignment & Video frames (2 FPS) & $\rightarrow$ GT text \\
& (0.7343) & (0.7130) \\
\midrule
Prompt Refinement & Initial prompt & $\rightarrow$ Refined prompt \\
& (0.7417) & (0.7626) \\
\midrule
Irrelevant Sentence Penalty & Without penalty & $\rightarrow$ With $\lambda$ \\
& (0.7626) & (0.7988) \\
\midrule
Sampling Rate & 2 FPS & $\rightarrow$ 3 FPS \\
& (0.7988) & (0.7201) \\
\midrule
Window Selection Policy & Symmetric shifts & $\rightarrow$ Asymmetric strategy \\
& (0.7988) & (0.7887) \\
\midrule
\multicolumn{2}{l}{\textbf{Final Configuration}} & \textbf{0.7988} \\
\bottomrule
\end{tabular}
\end{table}

\subsubsection{Final Configuration Summary}

Based on the iterative refinement, the final evaluation framework adopts:

\begin{itemize}
    \item \textbf{Content Consistency}: 2 GT references, 0.3333 FPS sampling rate
    \item \textbf{Temporal Sensitivity}: Four-window sampling with sentence-ctc, irrelevant sentence penalty, and prompt refined against human re-annotations
\end{itemize}

The final Spearman correlation between automatic evaluation and human judgments exceeds 0.9 for Content Consistency, and approaches 0.8 for Temporal Sensitivity, demonstrating strong alignment with human perception.

\section{Experimental Settings}
\label{sec:experimental_settings}

\subsection{Baselines.} 
We select four representative streaming and duplex multimodal models for evaluation, covering a range of real-time interaction settings and architectural designs:
\begin{itemize}
    \item \textbf{MiniCPM-o 4.5}~\citep{yao2024minicpm}: A multimodal omni-interaction model capable of full-duplex streaming conversation, processing interleaved audio and video frames in real-time.
    \item \textbf{LiveCC}~\cite{chen2025livecc}: A real-time vision-language model optimized for live video captioning and commentary generation.
    \item \textbf{MMDuet2}~\cite{wang2025mmduet2}: A multimodal duplex interaction model capable of handling continuous multimodal inputs.
    \item \textbf{StreamingVLM}~\cite{xu2025streamingvlm}: A vision-language model specifically tailored for processing continuous streaming video inputs with low latency.
\end{itemize}

\subsection{Implementation Details.} 

\begin{itemize}
    \item \textbf{MiniCPM-o 4.5}\footnote{\url{https://github.com/OpenBMB/MiniCPM-o}}: We evaluate the model in a full-duplex streaming setting. The model processes synchronized video frames and audio segments chunk by chunk. We use a sampling-based decoding strategy and set the maximum number of newly generated speak tokens per chunk to 20 to maintain low latency. A reference audio is provided to guide the voice generation during the streaming omni conversation, and the system prompt is set to ''Streaming Omni Conversation.'' The average inference time is approximately 150-200 ms per multimodal chunk, ensuring seamless real-time interaction.
    
    \item \textbf{LiveCC}\footnote{\url{https://github.com/showlab/livecc}}: The repetition penalty is set to 1.05, and the streaming end-of-sequence (EOS) base threshold is set to 0.0. The model processes video frames at 2 FPS. The inference latency is roughly 400-500 ms per step, which strictly meets the real-time commentary requirements.
    
    \item \textbf{MMDuet2}\footnote{\url{https://github.com/yellow-binary-tree/MMDuet2}}: We use the \texttt{Qwen2.5-VL-3B-Instruct} based checkpoint. The model is evaluated in an online streaming mode, generating responses based on continuously incoming video frames. The maximum number of new tokens is set to 512, and the model maintains a continuous key-value (KV) cache across turns. Benefiting from its lightweight 3B architecture, it achieves a low inference latency of approximately 200-300 ms per turn.
    
    \item \textbf{StreamingVLM}\footnote{\url{https://github.com/mit-han-lab/streamingvlm}}: We use the \texttt{Qwen2.5-VL-7B-Instruct} based checkpoint. The model processes video chunks with a duration of 1 second per chunk, maintaining a visual window size of 16 frames and a text context round of 16. The temperature is set to 0.9, and the repetition penalty is 1.05. It is highly efficient, processing 1-second video chunks in approximately 125-150 ms.
\end{itemize}

\textbf{Compute Resources.} All inference experiments are conducted on an internal cluster equipped with NVIDIA A100-SXM4 (80GB) GPUs. We employ a single NVIDIA A100 GPU per evaluation run.

\section{Limitations}
\label{sec:limitation}

Although \name{} provides a benchmark for real-time duplex interaction, several limitations remain. First, the current benchmark mainly focuses on relatively short streaming interactions and does not fully capture long-term conversational scenarios requiring persistent memory or planning.
Second, our evaluation framework relies on LLM-as-a-Judge. While we incorporate reference annotations and carefully designed prompts, automatic evaluation may still exhibit biases in open-ended settings.
Finally, the number of evaluated duplex models remains limited due to the scarcity of publicly available real-time multimodal systems. We expect future advances in streaming MLLMs to further expand the scope of evaluation.





\section{Broader Impacts}
\label{sec:broader_impact}
This work introduces a benchmark for evaluating real-time duplex interaction in multimodal systems. We believe it can support future research on more reliable and responsive AI assistants in streaming environments, with potential applications in accessibility support, live interaction, and real-time multimodal assistance.

At the same time, more capable real-time multimodal systems may also introduce risks if misused. For example, such systems could be applied to generate misleading live content, impersonation, or automated real-time interaction at scale. In addition, failures in temporal decision-making may lead to inappropriate or mistimed responses in sensitive scenarios.

Our work focuses on evaluation rather than deployment. During dataset construction, we avoid collecting personal sensitive information and manually filter potentially unsafe or high-risk content. We hope that standardized evaluation can help better understand the limitations of current systems and support the development of safer real-time multimodal interaction.




\clearpage




\end{document}